\documentclass{article-hermes}  

\usepackage[utf8x]{inputenc}
\usepackage[T1]{fontenc}
\usepackage{xcolor}
\usepackage{url}

\title[Short title]{Étude de l'informativité des transcriptions : une approche basée sur le résumé automatique}

\author{Carlos-Emiliano González-Gallardo\fup{*,**} \andauthor Malek Hajjem\fup{*} \andauthor Eric SanJuan\fup{*} \andauthor Juan-Manuel Torres-Moreno\fup{*,**}  }
\address{
\fup{*} LIA, Université d'Avignon et des Pays de Vaucluse\\
339 chemin des Meinajariès, Agroparc BP 91228, F-84911 Avignon cedex 9, France\\
\{carlos-emiliano.gonzalez-gallardo, malek.hajjem, eric.sanjuan, juan-manuel.torres\}@univ-avignon.fr \\
\fup{**} Département de Génie Informatique et Génie Logiciel, École Polytechnique de Montréal. 
2900 Blvd. Edouard-Montpetit, H3T 1J4 Montréal (Québec) Canada\\
}
\resume{Dans cet article nous proposons une nouvelle approche d’évaluation de l’informativité des transcriptions issues de différents systèmes de Reconnaissance Automatiques de la Parole.
Cette approche, fondée sur la notion d'informativité, s'inscrit particulièrement dans le cadre du Résumé Automatique de texte effectué sur ces transcriptions.
Nous estimons, dans un premier temps, le contenu informatif des différentes transcriptions. 
Par la suite, nous explorons la capacité du Résumé automatique de texte pour surmonter la perte informative.
Pour ce faire, nous utilisons un protocole d’évaluation automatique de résumés sans références basé sur le contenu informatif.
Le but est de calculer les divergences entre les distributions de probabilité des différentes représentations textuelles obtenues: des transcriptions manuelles et automatiques et leurs résumés.
À l’issue d’une série d’évaluations, cette analyse nous a permis de juger à la fois la qualité des transcriptions en terme d’informativité et d’évaluer la capacité du résumé automatique de texte à compenser les problèmes soulevés lors de la phase de transcription.} 

\abstract{
In this paper we propose a new approach to evaluate the informativeness of transcriptions coming from Automatic Speech Recognition systems.
This approach, based in the notion of informativeness, is focused on the framework of Automatic Text Summarization performed over these transcriptions.
At a first glance we estimate the informative content of the various automatic transcriptions, then we explore the capacity of Automatic Text Summarization to overcome the informative loss.
To do this we use an automatic summary evaluation protocol without reference (based on the informative content), which computes the divergence between probability distributions of different textual representations: manual and automatic transcriptions and their summaries. 
After a set of evaluations this analysis allowed us to judge both the quality of the transcriptions in terms of informativeness and to assess the ability of automatic text summarization to compensate the problems raised during the transcription phase.}
\motscles{
Mesures d’évaluation,
Résumé automatique de textes,
Transcription automatique, 
Métriques d'informativité
}


\begin{document}

\maketitlepage

\newcommand{\fakesentence}{Attention à ce que les figures et les tableaux ne débordent pas dans les marges. }
\newcommand{\fakeparagraph}{
\fakesentence
\fakesentence
\fakesentence
\fakesentence
\fakesentence
\fakesentence
}

\section{Introduction}

\noindent Le résumé automatique de la parole constitue un cas particulier du résumé automatique de documents (RAD) \cite{Torres2014}.
Le flot d'entrée dans cette tâche est un flot de parole continue et la sortie sera sous la forme d'un résumé écrit ou parlé.
L'une des méthodes la plus répandue de génération des résumés automatiques de la parole est de tirer parti des transcriptions automatiques du discours parlé \cite{ding2012beyond,szaszak2016summarization,taskiran2001automated}.
Ce discours peut venir évidemment d'une vidéo.
L'idée consiste donc à utiliser les algorithmes développés dans le cadre du résumé automatique de texte, afin de produire une version synthétique à partir de l'information contenue dans la parole présente dans les vidéos. 
Pour ce faire, une étape de transcription automatique du contenu parlé est nécessaire.
Toutefois, dans ce cas, les traitements ne peuvent plus compter sur une transcription parfaite du texte à résumer, et doivent être capables de gérer les erreurs produites lors de l'étape de transcription.
En effet, aux éventuelles erreurs du système de résumé automatique s'ajoutent les limites des systèmes de Reconnaissance Automatique de la Parole (RAP).
Par conséquent, prévoir une stratégie pour estimer à quel point les méthodes issues du domaine textuel sont influencées par un contenu audio, devient indispensable.
Ce travail de recherche s'inscrit dans ce cadre précis.

Nous visons l'évaluation de l'informativité d'un ensemble de résumés automatiques générés à partir des sorties des différents systèmes de transcription automatique.
Les principales motivations de cette analyse s'inspirent essentiellement du besoin accru d'une représentation synthétique des informations parlées les plus importantes.
En effet, de nos jours, les vidéos numériques représentent un facteur important pour véhiculer l'information.
Avec l'avènement de plusieurs sites web d'hébergement vidéo tels que YouTube, Dailymotion, Hulu et autres; l’utilisateur a la possibilité de regarder les émissions des chaînes TV à travers des podcasts.
Cette rediffusion des émissions conduit à l'explosion du nombre de documents disponibles, ce qui rend nécessaire la gestion efficace de ce contenu vidéo dont le volume ne cesse pas d'augmenter.

Pour y arriver, une méthode de résumé par extraction est appliquée au contenu parlé, transcrit préalablement de façon automatique.
Notre contribution concerne plus spécifiquement la phase d'évaluation de ces résumés obtenus.
Nous cherchons, d'une part, à explorer la capacité d'un résumeur automatique à compenser la perte d'information issue des erreurs de transcription.
D'autre part, nous cherchons à estimer l'influence du bruit généré par ces transcriptions sur les résumés automatiques. Ceci sera mesuré en termes du contenu informatif retenu à l'issue du processus d'extraction des passages les plus représentatifs.

La suite de l’article est organisée comme suit.
Dans la section \ref{sec:etat}, nous présentons les travaux connexes.
Dans la section \ref{sec:corpus}, nous présentons les données utilisées dans le cadre de l’évaluation que nous proposons.
La section \ref{sec:mesureinformativite} introduit une approche analytique qui vise à évaluer les résumés automatiques générés à base des transcriptions issues de différents systèmes RAP.
Nous mettrons l'accent sur la mesure de l'informativité à travers des analyses poussées dont l'objectif est d'estimer la qualité des transcriptions en terme du contenu informatif et de la capacité du RAD à compenser les erreurs de transcription, ainsi dans cette section nous discutons les résultats obtenus.
Enfin, dans la section \ref{sec:onclusion} nous présentons les conclusions de cet article.

\section{État de l'art}
\label{sec:etat}


\subsection{Reconnaissance automatique de la parole (RAP)}
\noindent La RAP est une démarche qui permet de passer d’un signal acoustique de parole à sa transcription dans une version écrite \cite{rabiner1993fundamentals,hatoninria00105908}.
Elle consiste à employer des processus d'appariement afin de comparer une onde sonore à un ensemble d'échantillons, composés généralement de mots mais aussi des unités sonores encore plus élémentaires appelées phonèmes \cite{deng2013machine}.

On distingue deux approches pour la reconnaissances de ces formes élémentaires. 
La première est à base de règles utilisant des formes primitives préalablement déterminées par des experts \cite{lee2009recent}.
La réussite des expériences repose sur des conditions très strictes : vocabulaire réduit, mots isolés, peu de locuteurs, enregistrements dans des conditions de laboratoire \cite{5212951}.
Ainsi, ces méthodes acoustiques seules sont insuffisantes et ont donné lieu à peu de réalisations concrètes car elles posent le problème de l’inférence des règles ainsi que celui de l’extraction des formes primitives \cite{hatoninria00105908}.
En conséquence, des informations linguistiques commencent à être prises en compte dans les systèmes de reconnaissance pour ajouter des mots de contexte aux systèmes et ainsi améliorer la performance de la reconnaissance.
Ceci a donné naissance à la deuxième approche qui consiste à extraire des vecteurs de paramètres caractéristiques à partir de ces formes afin d'utiliser une technique de classification permettant d’accorder une classe à une forme donnée \cite{1162650,1055384,1163259}.
Dans ce cadre, les systèmes de reconnaissance de parole utilisent une approche statistique dites décodage statistique de la parole décrit dans \cite{Jelinek1998SMS280484}.


Cette conversion parole-texte doit être indépendante de la taille du vocabulaire, de l'accent du locuteur, de son niveau de stress, etc.
En effet, afin d'obtenir une transcription correcte, le système de transcription doit être capable de gérer les spécificités de la parole.
Ainsi, il est évident que les performances des systèmes de RAP dépendent du type et de la qualité des données à transcrire \cite{galiberthal01083636}.
Généralement la performance d'un système de RAP est mesurée en termes de taux d'erreur de mots (\textit{Word Error Rate}, WER).
Le WER consiste à dénombrer les erreurs de transcription et à les normaliser par le nombre total de mots dans la référence pour fournir un pourcentage d’erreurs.
Une telle métrique semble être efficace lorsque la transcription automatique est une fin en elle même.
Cependant de nos jours un module de RAP est généralement combiné à plusieurs autres modules de Traitement Automatique du Langage naturel (TAL) afin de pouvoir résoudre des tâches encore plus complexes \cite{benjannethal01134868}, et le WER n'est plus adapté.

Le WER accorde le même poids à toutes les erreurs (erreurs d’insertion, d’omission ou de substitution).
Cette stratégie d’évaluation entrave la prise en considération du cadre applicatif final. 
Dans ce contexte, un certain nombre de mesures alternatives ont été proposées dans la littérature.
On cite la perte relative d'information (\textit{Relative Information Loss}, RIL) \cite{miller1955nbi}, une mesure qui propose d’évaluer la perte d’information causée par les erreurs des systèmes de RAP.
Cette métrique se base sur l'information mutuelle pour obtenir la force de la dépendance statistique entre le vocabulaire de la référence {\bf X} et les mots de l’hypothèse {\bf W}.
Une approximation du RIL, c'est la perte d'information des mots (\textit{Word Information Lost}, WIL) proposée par \cite{confinterspeechMorrisMG04}.
Cette métrique estime aussi la perte d’information due à des erreurs de transcription.
Contrairement à la RIL, WIL prend en compte les mots corrects et les substitutions au moment de comparaison entre la référence et l’hypothèse.
Une autre proposition de \cite{mccowanrr0473} consiste à reprendre les mesures de Recherche d'Information (RI) pour l'estimation de perte d'information causée par les dérives de la transcription.
\cite{benjannethal01134868} suggèrent une nouvelle méthodologie pour évaluer la qualité des transcriptions automatiques dans le contexte d’extraction d’entités nommées à partir de documents transcrits automatiquement.
Ainsi ces différents travaux de l’état de l'art confirment la nécessité de revoir l’évaluation des transcriptions automatiques à base de WER et d'explorer d'autres types de métriques mieux adaptées pour mettre en évidence la qualité des transcriptions automatiques en termes d'informativité.

\subsection{Les mesures d'informativité et domaines d'application}

\noindent L'évaluation de l'informativité par identification de pépites informationnelles a été proposée dans un premiers temps dans \cite{confnaaclNenkovaP04}.
Cette méthode est basée sur la notion de  \textit{Summary Content Units} (SCU) ou pépites (\textit{Nuggets}) définies manuellement par des annotateurs.
Ces unités informationnelles, auxquelles les annotateurs accordent des poids, correspondent sensiblement au même contenu exprimé différemment.
Un résumé automatique est dit informatif s'il est bien noté et qu'il contient des unités de fort poids.

Cette notion d'informativité a été étendue pour couvrir différentes briques technologiques issues de TAL.
On la retrouve dans la tâche de traduction automatique en utilisant la méthode BLEU (\textit{Bilingual Evaluation Understudy})  \cite{papineni2002bleu} et aussi sur l'évaluation des résumés automatiques en utilisant la méthode ROUGE (\textit{Recall-Oriented Understudy for Gisting Evaluation}) \cite{Lin2004}. Le principe général de ROUGE est de calculer l'intersection des $n$-grammes communs entre le résumé automatique à évaluer et les résumés de référence disponibles.
Pour une évaluation correcte avec ROUGE, les travaux de l'état de l'art ont montré qu'il est nécessaire de produire au moins cinq résumés de référence générés par différents annotateurs \cite{Louis2009PCE16090671609127}.
Ainsi, ROUGE s'avère inappropriée lorsque la génération des résumés de référence est trop coûteuse comme dans le cas de documents très longs ou d’un très grand nombre de documents à résumer, ce qui représente le cas typique de la RI.

En effet dans le cadre de la RI cette notion de l'informativité à été reprise par la divergence de Kullback-Leibler (KL) \cite{kullback1951information}. 
Cette mesure permet de comparer le contenu d'un résumé produit automatiquement à celui du document source.
La comparaison est principalement basée sur l'étude des distributions de mots ou de ensembles de mots entre le résumé et les documents \cite{101007978364214556835}.
Une telle comparaison à base de distributions de probabilité est peu sensible à la présence (ou l'absence) de séquence de mots communs entre le résumé et la référence.
Ceci la rend difficile à adapter pour les résumés guidés par une requête ou concernant un sujet particulier \cite{Bellot2015a}.
Ce problème se concrétise dans le cadre de la contextualisation de tweets \cite{101007978364223577124}, une tâche qui combine les notions de RI et de résumé automatique.


{\sc Fresa}  ({\it FRamework for Evaluating Summaries Automatically})\footnote{{\sc Fresa} est téléchargeable à l'adresse: \url{http://fresa.talne.eu}} est une méthode automatique inspirée des travaux de \cite{louis-nenkova2009EMNLP} et de \cite{lin06} pour évaluer les résumés sans utiliser des références qui a été introduite par  \cite{torres10poli,saggion10}.
La méthode intègre un prétraitement classique des documents (filtrage des mots non porteurs d'information, normalisation, etc.) avant de calculer la divergence des distributions de probabilités entre le document source et le résumé candidat.
Ce prétraitement des documents permet de garder seulement les mots porteurs d'information et se focaliser sur l'informativité.
Pour le calcul de la divergence $D(P||Q)$, {\sc Fresa} a la possibilité de calculer la divergence de Jensen-Shannon ($\mathcal{JS}$) et également une modification de KL \cite{Torres2014} au moyen d'uni-grammes ({\sc Fresa$_1$}), bi-grammes ({\sc Fresa$_2$}), bi-grammes-SU4 ({\sc Fresa$_4$}) et leur moyenne ({\sc Fresa$_M$}).

{\sc Fresa} a été utilisée pour évaluer la qualité des résumés de documents biomédicaux en langue catalane, où les résumés de référence des auteurs n'étaient pas disponibles \cite{vivalditerminalia10}. 
Egalement, {\sc Fresa} a été employé lors de la campagne INEX 2010\footnote{\url{http://www.inex.otago.ac.nz/tracks/qa/qa.asp}}, volet question-réponse (\textit{QA Track}, QA@INEX) \cite{torres10poli}. 
Cette dernière tâche combine les démarches de RI et du résumé automatique.


\section{Corpus} \label{sec:corpus}

\noindent Notre intérêt à l'analyse de l'informativité des transcriptions automatiques a été principalement motivé par l'explosion des données audiovisuels due à la rediffusion des émissions des chaînes TV à travers des podcasts d'actualités.
Pour une meilleure visibilité du concept d'informativité, nous avons pris en compte le contexte multilingue à travers une collection de vidéos en français et en anglais.

Pour chaque langue nous avons sélectionné 10 vidéos à partir de l'hébergeur web YouTube pour les chaînes d'actualités : France24, RT, Euronews et BBC.
Les documents audiovisuels de ce corpus ont été collectés par \cite{leszczuk2017video}.
Différentes thématiques ont été abordées dans le but de minimiser la prépondérance d'un thème sur la totalité des sujets abordés. 
Le tableau \ref{table:topics} illustre la distribution des thèmes du corpus.

\begin{table}
  \centering
  \caption{Thèmes des vidéos}  
  \begin{tabular}{cc}
  \hline
  Thème & Nombre de vidéos \\
  \hline
  Syrie & 6 \\
  Territoires occupés & 4 \\
  Donald Trump & 3 \\
  Droits de l'homme & 3 \\  
  Terrorisme & 3 \\
  Technologie & 1 \\
  \hline
\end{tabular}
\label{table:topics}
\end{table}



En ce qui concerne la création du corpus textuel, une étape de transcription automatique du contenu parlé a été appliquée à travers de trois systèmes de RAP.
En plus des sorties de ces trois systèmes, nous avons produit une transcription manuelle (dite transcription de référence, Réf-humaine) faite par différents experts maîtrisant la langue de la vidéo, afin d'avoir une ressource pour évaluer les performances des systèmes RAP.
Pour une étude plus complète, nous avons décidé de tirer parti à la fois des systèmes RAP commerciaux et académiques. 

La transcription automatique à base d'un système non commercial a été réalisée en utilisant KATS (\textit{Kaldi-Based Transcription System}), système RAP qui a été introduit dans \cite{fohr2017new} et qui utilise des modèles acoustiques à base des réseaux de neurones profonds.
En ce qui concerne les systèmes commerciaux, nous avons utilisé le système Google Cloud Speech API\footnote{\url{https://cloud.google.com/speech}} (Google-ASR) ainsi que le système IBM speech-to text\footnote{\url{https://www.ibm.com/watson/services/speech-to-
ext}} (IBM-ASR)  \cite{saon2015ibm}.
Ces deux derniers systèmes utilisent des modèles à base des réseaux de neurones combinés à d'autres statistiques.

La description du corpus de transcriptions (français et anglais) en termes de vocabulaire a été synthétisée au tableau \ref{table:wordspertranscription}.
En effet, les vidéos les plus courtes impliquent un contenu textuel inférieur ou égale à 240 mots.
En revanche les vidéos les plus longues excédent les 2 350 mots.  

\begin{table}
  \centering
  \caption{Statistiques des transcriptions}  
  \begin{tabular}{llcc}
  \hline
  Langue & Système & Moyenne de mots & Ecart-type (mots) \\
  \hline
  & KATS & 950 & 704 \\
  Français &Google-ASR & 847 & 644 \\
  &IBM-ASR & 924 & 672 \\
  &Réf-humaine & 958 & 702 \\
  \hline
  & KATS & 846 & 480 \\
  Anglais &Google-ASR & 705 & 410 \\
  &IBM-ASR & 870 & 510 \\
  &Réf-humaine & 805 & 495 \\
  \hline  
\end{tabular}
\label{table:wordspertranscription}
\end{table}

\section{Mesure de l'informativité des transcriptions automatiques et l'impact du résumé automatique}
\label{sec:mesureinformativite}

\noindent Notre hypothèse est que le résumé automatique représente un moyen extrinsèque assez objectif pour évaluer la qualité des transcriptions venant d'un système RAP.
On sait que l'informativité contenue dans un résumé vis-à-vis la source est un bon indicateur de la qualité d'un système de résumé automatique \cite{confnaaclNenkovaP04,saggion10}.
Il est donc possible d'évaluer la qualité d'une transcription via la mesure d'informativité contenue dans le résumé correspondant.


Dans le cadre du résumé automatique, l'existence des phrases dans le texte source est essentielle pour repérer les phrases contenant les informations pertinentes.
Les transcriptions issues des différents systèmes RAP ne contiennent pas des signes de ponctuation et représentent une séquence continue de mots. 
Ainsi, une étape de segmentation est nécessaire et représente un enjeu particulier en soi.
Segmenter une transcription revient à établir des hypothèses des frontières de phrases en positionnant des signes de ponctuation dans la séquence initiale. 

\cite{gonzalez2018sentence} se sont intéressés en particulier à trouver les marqueurs de fin de phrase qui représentent les frontières des phrases en français.
Nous nous sommes inspirés de ce travail pour la génération automatique des segments venant des transcriptions.
Notre approche a été étendue dans le but de traiter la langue anglaise en plus du français.
L'architecture à base de réseaux de neurones de convolution que nous avons appliqué sur les transcriptions en anglais et français est la même architecture qui \cite{gonzalez2018sentence} ont signalé être la meilleure pour la segmentation des phrases en français.

Pour l'anglais et durant la phase d'apprentissage du réseau, nous avons utilisé un corpus de 426 millions de mots extraits de \textit{English Gigaword Fifth Edition}\footnote{\url{https://catalog.ldc.upenn.edu/LDC2011T07}}. 
L'évaluation du système a été appliquée sur un sous-ensemble de 106 millions de mots.
Concernant le français, nous avons utilisé un corpus de 470 millions de mots extraits de \textit{French Gigaword First Edition}\footnote{\url{https://catalog.ldc.upenn.edu/LDC2006T17}} et un sous-corpus de 117 millions de mots pendant la phase d'évaluation. 
Le tableau \ref{table:segmentsstats} illustre la performance des deux systèmes en termes de précision, rappel et Fscore (F1) pour prédire les fins de phrases.
Cette même stratégie de segmentation a été exécutée aussi bien pour les transcriptions automatiques que sur les transcriptions manuelles. 

\begin{table}
  \centering
  \caption{Performance des systèmes de segmentation des phrases}  
  \begin{tabular}{cccc}
  \hline
  Langue & Précision & Rappel & F1   \\
  \hline
  Français & 0,845 & 0,754 & 0,795\\
  Anglais & 0,838 & 0,796 & 0,816\\
  \hline  
\end{tabular}
\label{table:segmentsstats}
\end{table}


En ce qui concerne le résumé automatique, nous avons opté pour l'approche extractive, car en outre sa facilité d'implémentation, ses performances ont été bien établies dans les travaux de l'état de l'art \cite{Torres2014}. 
Cette approche consiste à extraire, parmi les $P$ phrases qui constituent un document source, les $n$ phrases portant la plus grande quantité d'information. Ces phrases sont censées être les plus relevantes pour produire un résumé.

\subsection{ {\sc Artex}}

\noindent {\sc Artex} (Autre Résumer de TEXtes) est un système de résumé automatique de documents par extraction de segments pertinents qui a été introduit par \cite{torres2012artex}.
La première phase d'{\sc Artex} consiste à faire un prétraitement du texte source. 
Il s'agit d'une étape très importante qui permet de normaliser les mots et de supprimer les mots outils (peu informatifs).
Ceci avec le but de réduire la dimensionnalité de la représentation vectorielle et de pouvoir calculer l'informativité des phrases. 
Une fois le texte prétraité, le calcul de deux vecteurs type centroïde est réalisé: un vecteur lexical moyen $VL_j, j=1...N$ (qui représente l'informativité moyenne du lexique de $N$ termes) et un vecteur thématique moyen $VT_i, i=1...P$ (qui représente le thème central du document composé de $P$ phrases).
Un produit scalaire de ces deux vecteurs (normalisé) est effectué pour chaque phrase $i$ du document du $i=1...P$.
Un poids pour chaque phrase sera donc obtenu moyennant une fonction à partir du produit scalaire.
À l'issue de la pondération des phrases, la génération du résumé est simple: il consiste à concaténer les phrases ayant les scores les plus élevés dans l’ordre de leur occurrence dans le texte source.
Une procédure de post-traitement (diminution de la redondance ou simplification) peut être appliquée à la fin du processus.

Dans nos expériences, le ratio de compression\footnote{Le ratio de compression représente le rapport entre la taille du résumé et la taille du document source (en nombre de phrases): $\rho=\frac{|\textrm{résumé}|}{|\textrm{source}|}$.} a été fixé à $\rho=0.35$ pour obtenir les résumés automatiques.
Le choix d'{\sc Artex} pour la génération des résumé des transcriptions automatiques et manuelles est justifié par sa simplicité d'implémentation, sa rapidité d'exécution et ses résultats compétitifs \cite{morchid2017automatic}.
Nous notons aussi que {\sc Artex} est assez indépendant des connaissances linguistiques, ce qui le rend particulièrement adapté à la proposition de résumer des transcriptions de documents audio en plusieurs langues\footnote{Bien sûr, d'autres systèmes de résumé automatique statistiques ou non auraient pu été utilisés dans cette tâche. 
Il s'agit d'un module du type \textsl{plug-in}.}.

\subsection{Évaluation de l'informativité}

\begin{figure}[h!]
\begin{center}
\frame{\includegraphics[trim={0cm 0cm 0cm 0cm},clip=true,width=1.0\textwidth]{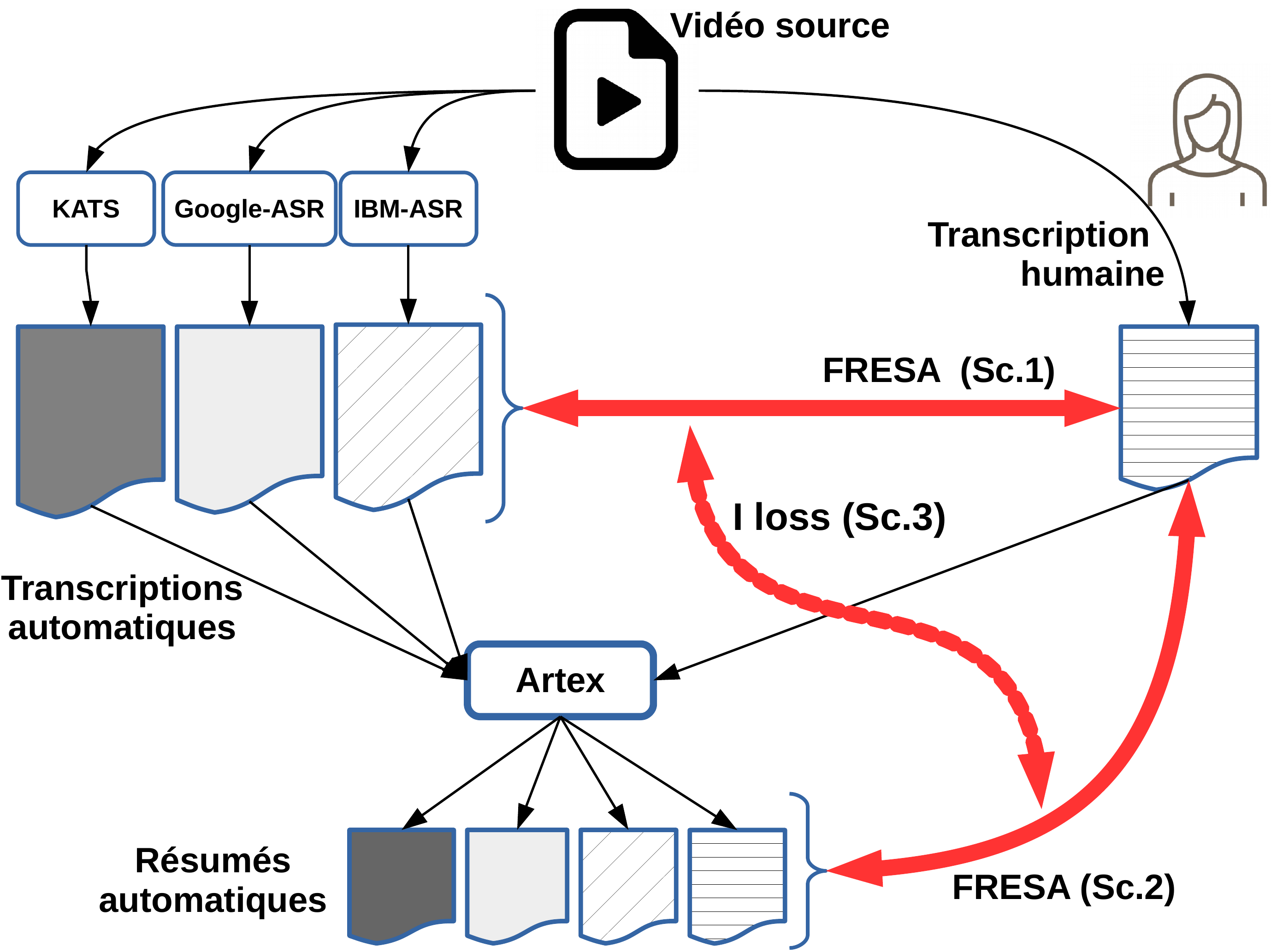}}
\caption{\label{fig:informativeness}
Protocole d'évaluation utilisé lors de nos expériences. Scénarios d'évaluation : Sc.1 Scénario 1, Sc.2 Scénario 2 et Sc.3 Scénario 3.}
\end{center}
\end{figure}

La figure \ref{fig:informativeness} illustre le protocole que nous avons suivi pour évaluer, d'abord l'informativité des transcriptions automatiques et ensuite l'impact du résumé automatique sur l'informativité.

Tout d'abord, les transcriptions manuelles et automatiques ont été effectuées sur les vidéos comme décrit dans la section \ref{sec:corpus}.
Ensuite, des résumés automatiques ont été générés en utilisant {\sc Artex}.
Enfin, pour mesurer l'informativité, nous avons calculé la divergence $\mathcal{KL}$ modifiée \cite{Torres2014} entre les distributions de probabilité des transcriptions manuelles et automatiques, ainsi que la divergence entre les distributions de probabilités des transcriptions manuelles et des résumés automatiques en utilisant la méthode {\sc Fresa}.



Nous proposons trois scénarios d'évaluation basés sur les scores {\sc Fresa}.
Dans le premier scénario (Sc.1), nous comparons l'informativité entre la transcription humaine et les transcriptions automatiques produites par différents systèmes RAP.
Puis, dans un deuxième scénario (Sc.2), l'informativité est mesurée entre la transcription humaine et les résumés produits à partir des transcriptions automatiques et manuelles. L'application de {\sc Fresa} entre la référence humaine et son résumé établit une valeur maximale d'informativité attendue, qui est comparée à l'informativité des résumés venant des transcriptions automatiques.
Enfin, pour un troisième scénario (Sc.3), nous comparons un ratio d'informativité de Sc.1 par rapport à Sc.2 afin d'évaluer la capacité du résumé à surmonter la perte d'informativité des systèmes RAP.
Les tables \ref{table:tm_vs_ta_French} à \ref{table:tm_vs_sums_English} affichent les valeurs moyennes et l'écart-type des mesures {\sc Fresa}.
Nous avons considéré important de prendre en compte l'écart-type car cette valeur donne une idée générale de la façon dont l'informativité est influencée par les variations de longueur et de diversité des sujets traités.

\subsection{Résultats}

\begin{itemize}

\item{Transcription manuelle vs. Transcriptions automatiques (Sc.1)}


\noindent Les scores {\sc Fresa} pour le français et l’anglais concernant le premier scénario sont montrés dans les tableaux \ref{table:tm_vs_ta_French} et \ref{table:tm_vs_ta_English} respectivement. En français, on peut observer que le système KATS maintient un degré d’informativité supérieur sur toutes les mesures {\sc Fresa}. 
L’informativité la plus basse est produite par le système IBM-ASR avec un score moyen ({\sc Fresa}$_M$) de 0,539.


Un comportement différent est observé pour l’anglais. 
Google-ASR obtient la meilleure informativité sur presque tous les scores {\sc Fresa}. Ceci est probablement du au fait que l'algorithme Google-ASR utilise plus de ressources pour l'anglais que pour le français.
Un détail intéressant à signaler concerne le système IBM-ASR. Ce système produit le plus bas écart-type sur tous les scores {\sc Fresa}, ce qui suggère qu’un léger mais stable degré d’informativité a été partagé sur les différents thèmes et les différentes longueurs.

\newpage

\begin{table}
  \centering
  \caption{Sc.1 - Transcription manuelle vs. Transcriptions automatiques (français)}  
  \begin{tabular}{ccccc}
  \hline
  SYSTEME & {\sc Fresa}$_1$ & {\sc Fresa}$_2$ & {\sc Fresa}$_4$ & {\sc Fresa}$_M$  \\
  \hline
   KATS & \textbf{0,835} $\pm$ \textbf{0,076}& \textbf{0,697}$\pm$ \textbf{0,118 }& \textbf{0,683 }$\pm$ \textbf{0,130 }& \textbf{0,738}$\pm$ \textbf{0,106}\\
   Google-ASR & 0,795 $\pm$ 0,132& 0,664 $\pm$ 0,145&0,659 $\pm$0,148 & 0,706$\pm$  0,140\\
   IBM-ASR & 0,662 $\pm$ 0,141& 0,485 $\pm$ 0,134 & 0,471 $\pm$ 0,141  & 0,539 $\pm$0,137\\

  \hline  
\end{tabular}
\label{table:tm_vs_ta_French}
\end{table}

\begin{table}
  \centering
  \caption{Sc.1 - Transcription manuelle vs. Transcriptions automatiques (anglais)}  
  \begin{tabular}{ccccc}
  \hline
  SYSTEME & {\sc Fresa}$_1$ & {\sc Fresa}$_2$ & {\sc Fresa}$_4$ & {\sc Fresa}$_M$  \\
  \hline
   KATS & \textbf{0,741  $\pm$ 0,061}&0,584  $\pm$ 0,089 &0,567 $\pm$ 0,094 & 0,631$\pm$0,080\\
   Google-ASR & 0,740 $\pm$ 0,093& \textbf{0,605 $\pm$ 0,137} & \textbf{0,590$\pm$0,139}&\textbf{0,645 $\pm$0,122}\\
   IBM-ASR & 0,736 $\pm$ 0,058& 0,578 $\pm$0,076  & 0,566$\pm$ 0,082 & 0,626$\pm$0,070\\

  \hline  
\end{tabular}
\label{table:tm_vs_ta_English}
\end{table}

\item{Transcription manuelle vs. Résumés des transcriptions (Sc.2)}


\noindent Les tableaux \ref{table:tm_vs_sums_French} et \ref{table:tm_vs_sums_English} montrent les résultats pour le deuxième scénario.
La valeur maximale d’informativité attendue est établie par les scores basés sur les références humaines, et correspond au plus grand score {\sc Fresa} qu’un résumé peut obtenir vis-à-vis de la référence manuelle.
Plus le score {\sc Fresa} du résumé automatique est proche de cette valeur, plus il sera informatif.
Pour le français, Google-ASR possède le score moyen le plus proche ($0,286$) à la valeur d'informativité maximale attendue ($0,395$). Par contre, pour l'anglais, le score moyen d'informativité $0,303$ a été obtenu par KATS, c'est-à-dire $0,003$ plus haut qu'IBM-ASR.

Dans le tableau \ref{table:wordspertranscription} on peut voir que Google-ASR a produit le plus bas nombre moyen de mots par transcription pour le français et l'anglais. Après une analyse manuelle de ces transcriptions, nous avons observé que lorsqu'une partie de l'audio dans la vidéo est difficile à comprendre, Google-ASR ne génère pas la transcription correspondante. 
Cela semble produire des effets opposés en fonction de la langue. Le processus de résumé a un impact positif pour le français, en éliminant de manière efficace les parties les moins informatives de la transcription. 
Ce comportement semble être à l'opposé en anglais, où l'informativité de Google-ASR passe de la première position de Sc.1 à la dernière de Sc.2.
Nous pensons que l'excès moyen de mots transcrits que KATS et IBM-ASR génèrent par rapport à la transcription manuelle (tableau \ref{table:wordspertranscription}) influencent le score d'informativité de ces deux systèmes après le résumé.

\begin{table}[h!]
  \centering
  \caption{Sc.2 - Transcription manuelle vs. Résumés (français)}  
  \begin{tabular}{ccccc}
  \hline
  SYSTEME & {\sc Fresa}$_1$ & {\sc Fresa}$_2$ & {\sc Fresa}$_4$ & {\sc Fresa}$_M$  \\
  \hline
   KATS & \textbf{0,385 $\pm$0,080} & 0,238$\pm$ 0,067 & 0,213 $\pm$ 0,069&0,279 $\pm$ 0,069\\
   Google-ASR &0,377  $\pm$0,098 &\textbf{0,249 $\pm$ 0,083} &\textbf{ 0,231 $\pm$0,087} &\textbf{0,286 $\pm$0,087}\\
   IBM-ASR & 0,352 $\pm$ 0,076&0,200 $\pm$ 0,071 & 0,181 $\pm$0,069& 0,244$\pm$0,069\\
   \hline  
   Réf-humaine & 0,461 $\pm$0,065 &0,371 $\pm$0,047  & 0,352 $\pm$0,049 &0,395 $\pm$0,051\\
  \hline  
\end{tabular}
\label{table:tm_vs_sums_French}
\end{table}

\begin{table}[h!]
  \centering
  \caption{Sc.2 - Transcription manuelle vs. Résumés (anglais)}  
  \begin{tabular}{ccccc}
  \hline
  SYSTEME & {\sc Fresa}$_1$ & {\sc Fresa}$_2$ & {\sc Fresa}$_4$ & {\sc Fresa}$_M$  \\
  \hline
   KATS & 0,395 $\pm$ 0,078& \textbf{0,266}$\pm$ \textbf{0,068} &\textbf{0,248 }$\pm$\textbf{0,063} &\textbf{0,303} $\pm$\textbf{0,067}\\
   Google-ASR &0,342  $\pm$0,085 &0,222 $\pm$ 0,076 & 0,202 $\pm$ 0,069&0,256 $\pm$0,075\\
   
   IBM-ASR & \textbf{0,396 $\pm$ 0,079}&0,261 $\pm$0,069  & 0,242 $\pm$0,067 & 0,300$\pm$0,070\\
   \hline  
   Réf-humaine & 0,441 $\pm$ 0,042&0,347 $\pm$ 0,033 &0,325  $\pm$0,032 &0,371 $\pm$0,034\\

  \hline  
\end{tabular}
\label{table:tm_vs_sums_English}
\end{table}

\item{Résumé automatique vs. Perte d'informativité (Sc.3)}


\noindent Le scénario Sc.3 prend en compte les scores du scénario Sc.1 et du scénario Sc.2 afin de calculer la perte d’informativité produite par le résumé automatique. 
La perte d’informativité est exprimée comme un ratio entre le score {\sc Fresa} de résumés venant de transcriptions automatiques et manuelles.
Elle est donnée par l’équation [\ref{formula:I_loss1}].

\begin{equation}
  I_\textrm{loss}=100\times \left( 1- \frac{ \textrm{\sc Fresa}_{x_\textrm{système}} } {\textrm{\sc Fresa}_{x_\textrm{Réf-humaine}}} \right)
\label{formula:I_loss1}
\end{equation}

\noindent où ${\textrm{\sc Fresa}_{x_\textrm{Réf-humaine}}}$ est égale à $1$ pour le scénario Sc.1.


La figure \ref{fig:lossFrenchEnglish} montre la perte d’informativité {\sc Fresa}$_M$ en français et en anglais. On peut observer dans le cas du français, que les systèmes Google-ASR et IBM-ASR ont une perte plus petite après que le résumé automatique a été généré, avec une $I_{loss_{Sc.2}}$ qui diminue 1,8\% et 7,94\% respectivement. Par contre, le résumé automatique produit un impact négatif sur KATS, avec une augmentation de $I_{loss}$ de 3,27\%. Un comportement distinct se présente en anglais. 
La figure \ref{fig:lossFrenchEnglish} montre une augmentation de la performance en termes d'informativité. La perte de KATS tombe $18,61\%$, suivie par le système IBM-ASR ($18,15\%$) et Google-ASR avec $4,41\%$.

\begin{figure}[h!]
\begin{center}
\frame{\includegraphics[trim={0cm 0cm 0cm 0cm},clip=true,width=1\textwidth]{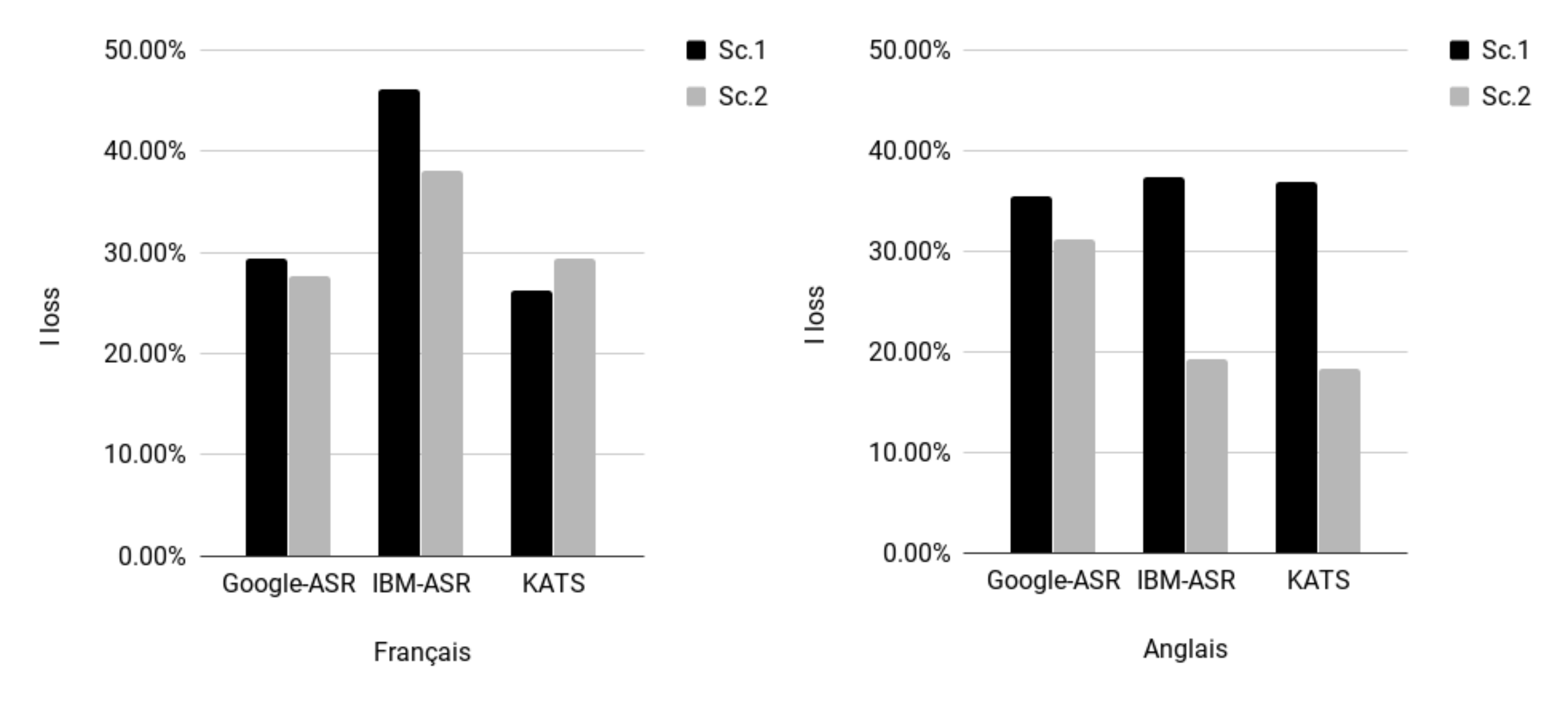}}
\caption{\label{fig:lossFrenchEnglish}Perte d’informativité pour l'anglais et français.}
\end{center}
\end{figure}

\end{itemize}

\section{Conclusion et perspectives}
\label{sec:onclusion}

\noindent Dans cet article nous avons proposé une évaluation indirecte de la qualité des transcriptions au moyen d'une mesure de l'informativité dans le résumé automatique. 
Elle a été inspirée des travaux sur la recherche d'information ainsi comme du résumé automatique.
Ce type de méthode d'évaluation produit une meilleure compréhension de la qualité des transcriptions automatiques en comparaison avec les mesures standards comme WER.
À notre connaissance, il s'agit du premier travail qui introduit une telle mesure d'informativité dans l'évaluation de la qualité des transcriptions et de leur utilisation dans une chaîne de traitement automatique de langues. 

Nous avons montré qu'en général, le résumé automatique peut augmenter le contenu informatif venant des transcriptions automatiques.
Les résultats sont exploitables pour guider l'évaluation automatique (ou semi-automatique) des systèmes RAP couplés aux systèmes TAL.
La démarche reste indépendante du domaine des documents et assez indépendante de la langue, bien que nos tests aient seulement été réalisés pour le moment en français et en anglais. 
Des tests en langue arabe sont actuellement en cours.
En outre, nous allons augmenter le nombre de transcriptions manuelles afin de réduire le biais produit par leur nombre limité et également tester d'autres approches de résumé automatique.

\acknowledgements{Nous remercions le soutien financier du programme européen Chist-Era à travers du projet \textit{Access Multilingual Information opinionS (AMIS)}, ANR-France / Europe.}

\bibliography{coria}

\end{document}